# A new Stack Autoencoder: Neighbouring Sample Envelope Embedded Stack Autoencoder Ensemble Model

Chuanyan Zhou[1], Jie Ma[1], Fan Li[1], Yongming Li[1*], Pin Wang[1], Xiaoheng Zhang[1]

(1. School of Microelectronics and Communication Engineering, Chongqing University, Chongqing 400044, P.R. China)

**Abstract**

Stack autoencoder (SAE), as a representative deep network, has unique and excellent performance in feature learning, and has received extensive attention from researchers. However, existing deep SAEs focus on original samples without considering the hierarchical structural information between samples. To address this limitation, this paper proposes a new SAE model-neighbouring envelope embedded stack autoencoder ensemble (NE_ESAE). Firstly, the neighbouring sample envelope learning mechanism (NSELM) is proposed for preprocessing of input of SAE. NSELM constructs sample pairs by combining neighbouring samples. Besides, the NSELM constructs a multilayer sample spaces by multilayer iterative mean clustering, which considers the similar samples and generates layers of envelope samples with hierarchical structural information. Second, an embedded stack autoencoder (ESAE) is proposed and trained in each layer of sample space to consider the original samples during training and in the network structure, thereby better finding the relationship between original feature samples and deep feature samples. Third, feature reduction and base classifiers are conducted on the layers of envelope samples respectively, and output classification results of every layer of samples. Finally, the classification results of the layers of envelope sample space are fused through the ensemble mechanism. In the experimental section, the proposed algorithm is validated with over ten representative public datasets. The results show that our method significantly has better performance than existing traditional feature learning methods and the representative deep autoencoders.

## 1. Introduction

Feature learning is critical for machine learning, and the effectiveness of feature learning heavily affects the performance of machine learning algorithms [1] . In recent years, feature learning algorithms have received extensive attention and research by scholars at home and abroad [2-4]. Traditional feature learning algorithms mainly include least absolute shrinkage and selection operator (LASSO) [5], Relief [6], P_value[7], principal component analysis (PCA) [8], linear discriminant analysis (LDA) [9], locality preserving projections (LPP) [10], etc. Traditional feature learning algorithms have been widely used in applications such as machine vision and pattern recognition and achieved good results. However, traditional feature learning algorithms are based on shallow empirical knowledge and cannot effectively mine the complex nonlinear relationship between data, so they have certain limitations. During recent years, the boom in deep learning has led many researchers to try to use deep networks for feature learning. And it has been implemented in the fields of image classification [13]-[15], speech recognition [16], machine translation [17] and time series prediction [18]. Typical deep neural networks mainly include convolutional neural network (CNN) [15], recurrent neural network (RNN)[18]- [20], stack autoencoder (SAE) [20], deep belief network (DBN) [21], long short term memory (LSTM) [22] , etc. Compared with traditional feature learning, the main advantage of deep learning is its powerful feature self-learning ability. Through multilayer nonlinear transformation, the network can learn the main driving variables and main structural information in the input data, and obtain more essential information about the original data [23].

Among deep feature learning, SAE as an unsupervised learning algorithm can automatically learn effective features from a large amount of unlabeled data. It has its unique and excellent performance in feature classification, and has received extensive attention from researchers. Therefore，various networks have also been proposed by researchers successively. Rui Li *et al*. [24] proposed a supervised autoencoder, adding an additional classification layer on top of the representation layer to jointly predict the target and reconstruct the input, and then build a stacked supervised autoencoder for the classification task, the model is able to learn to recognize features, which significantly improves the classification performance. In order to solve the problem of finding feature representations that minimize the distance between source and target domains and data redundancy may lead to the performance degradation of transfer learning, Yi Zhu *et al*. [25] proposed a new deep learning framework for semi-supervised representations for transfer learning. To solve the problem of imbalanced learning, Nima Farajian *et al*. [26] proposed an imbalanced learning method that includes feature learning and classification steps. In the feature learning step, by stacking two regularized autoencoders, a one-class learning method is used to extract meaningful features from a small number of data and capture its underlying manifold, the proposed method achieves quite good performance. Ahmad M. Karim [27] integrates the post-processing procedure into the data classification framework and proposes a new data classification framework that combines a post-processing system composed of sparse autoencoders (SAEs) and a linear system model based on particle swarm optimization (PSO). The framework can be applied to any data classification problem with only minor updates, such as changing some parameters, including input features, hidden neurons, and output classes. Wenjuan Wang *et al*. [28] proposed an efficient Stacked Shrinkage Autoencoder (SCAE) method for unsupervised feature extraction in view of the large-scale, high-dimensional, and highly redundant network traffic characteristics in cloud computing environments. By using SCAE methods, better and more robust low-dimensional features can be learned automatically from raw network traffic. Weifeng Liu *et al*. [29] proposed a large-capacity autoencoder (LMAE) to further improve the discriminability by enforcing a large marginal distribution of samples of different classes in the hidden feature space. Stack single-layer LMAEs to build a deep neural network to learn suitable features. Finally, these features are put into a softmax classifier for classification. Due to the unsupervised self-construction of SAE in the pre-training stage, the correlation of deep features with fault types cannot be ensured. Wang Y *et al*. [30] proposed a cascaded supervised autoencoder to pretrain deep networks and obtain deep fault-related features from raw input data. In each supervised autoencoder, informative features are learned from the input data with the aim of distinguishing different failure types to a large extent. The network gradually learns high-order fault-related features from the original input data, which improves the classification accuracy of the classifier.

Although SAEs have achieved great success in many applications [31]-[33], it is still a challenging problem to use a deep autoencoder for effective feature learning [34]. Existing SAEs mainly considers minimizing the error between each input sample and its output with reconstructed deep features, so the structural information between samples is not mined. As shown in Fig.1, existing autoencoder do not consider the relationship (structure information) between the ith sample and the jth sample. However, ignoring the sample structure information will lead to the decrease of separability between samples, limit the search for optimal samples, and affect the classification performance of the algorithm [35]. So, it is necessary to consider the structure information among samples.

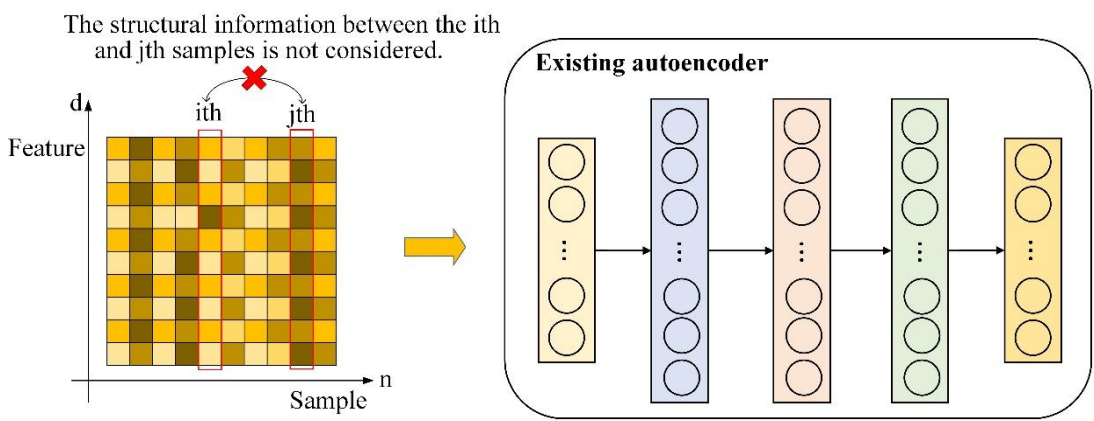

**Fig. 1** The existing autoencoder model.

Besides, it is challenging to find the optimal features for original samples. The Fig. 2 shows the distribution of original samples. As shown in Fig.2 (a), under feature space of F1 and F2, it is difficult to achieve their linear separability and to find a reliable optimal feature set. As shown in Fig.2 (b), the reparability is significantly improved after sample transformation. Sample transformation is helpful to mine the structural information between samples and construct 'bigger' samples with better separability. In addition, although SAE can conduct batch training on samples and obtain the structural information between samples indirectly, the number of samples in each batch should not be too large, otherwise the complexity of the encoder will be seriously increased. However, if it is too small, it cannot reflect the structure information between samples. Even if the whole sample can be input into SAE for training, the number of possible combinations will explode, and the risk of over fitting will increase significantly. For example, the number of features is $d$, and the combination of features is recorded as $C1(d)$, the number of samples is $n$, and the combination of samples is recorded as $C2(n)$, then the number of combinations of the two will be $C1(d) \times C2(n)$, the risk of overfitting will grow significantly. Therefore, it is important to combine the sample transformation and SAE, to consider the structure information among samples well.

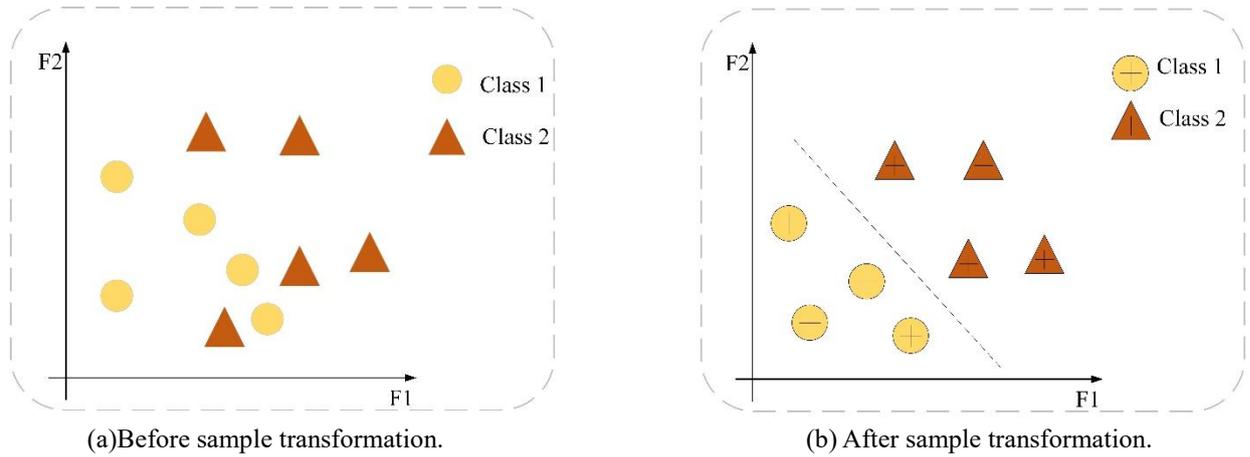

(a) Before sample transformation.  (b) After sample transformation.

**Fig. 2** Separability before and after reconstructing the sample.

The existing sample transformation methods are mainly include the method based on nearest neighbor rule [36], the method based on density [37], the method based on random sampling [38] and the method based on clustering [39] and so on. However, the nearest neighbor-based and density-based methods have very high computational complexity. The random sampling-based methods generally only obtain local optimal values and depend on the selection of the initial data set. The above three methods ignore the possible structural relationships between samples in the sample space. The clustering-based methods can separate a finite unlabeled data set into a finite and discrete set of data structures [41], analyzes the internal relationships of the data and aggregates similar samples together, so that the classifier focuses on samples belonging to different categories in the same cluster, thereby reducing the difficulty of classification facing a confusing boundary [42]. Clustering can be considered to mine sample structure information, and use the cluster center generated by sample clustering as a new sample to reconstruct the sample space. Existing clustering methods mainly include k-means algorithm [43], hierarchical clustering [44], density-based clustering [45], grid-based clustering[46], self-organizing map (SOM) [47] *et al*. Among them, k-means clustering can prune the tree according to the category of the lesser-known clustering samples, so as to determine the classification of some samples, which has the advantages of simplicity, easy implementation, fast convergence speed, excellent clustering effect, and strong interpretability of the algorithm. So, we consider using mean clustering to mine the hierarchical structure of the sample space, and construct multilayer sample spaces to obtain effective structural information. However, there will be differences in the distribution of samples between the multilayer sample spaces generated by iterative clustering [48], so we consider using the Maximum Mean Difference (MMD) to reduce the distribution difference of samples before and after clustering. MMD has been widely used in domain adaptation [49]. To measure differences in distributions and can be used to impose constraints on training models, optimized by updating training parameters [50].

In additional, complementarity between samples with original features and with deep features is important. However, existing SAEs did not consider this point.

To solve the problems above, this paper proposes a new SAE which consider structure information among samples well. The proposed SAE is called 'Neighboring Envelope Embedded Stack Autoencoder Ensemble'. First, the neighboring sample envelope learning mechanism (NSELM) is designed to preprocess the original samples and to construct layers of envelope samples for input of SAE. Second, the embedded stack autoencoder (ESAE) is designed to mine relationship between samples with original features and samples with deep features during training and within network. Third, several base classifiers are conducted on the layers of samples respectively. Fourth, multilayer space ensemble mechanism (MSEM) is designed to fuse the results from layers of classification. The main contribution of this paper can be expressed as follows.

(1) The neighboring sample envelope learning mechanism (NSELM) is proposed in this paper to construct hierarchical new samples (called 'envelope samples'). First, sample pair connection (SPC) is proposed to construct sample pairs. Then, interlayer consistency mechanism based on multilayer iterative mean clustering (ICMC) is used to construct the multilayer sample spaces to mine sample's structural information and achieve deep sample transformation. This mechanism deeply mines the information among the samples, thereby constructing more powerful 'bigger' samples (called envelope samples).

(2) SPC is proposed to mine relationship between neighboring samples, thereby considering structural information among samples and obtaining 'bigger' samples with more abundant features.

(3) ICMC is proposed to mine relationship between similar samples, thereby considering structural information among samples and obtaining more powerful samples.

(4) The existing stack autoencoder learning features have poor quality, insufficient complementarity with the original features, and limited feature complementarity fusion performance. The embedded stack autoencoder (ESAE) is designed in this paper to mine relationship between samples with original features and samples with deep features during training and within network. The ESAE is able to achieve high-quality deep features.

(5) Based on the NSELM and ESAE, a new SAE is proposed and it is called neighboring envelope embedded stack autoencoder ensemble (NE_ESAE). This model realizes the sample-feature cooperative transformation, while existing deep neural network only realize the feature transformation.

The rest of this paper is organized as follows. Section 2 discusses related work. Section 3 mainly describes the proposed method. Section 4 presents the experiments and results. The last section is the discussion and conclusions.

## 2. Related works

This section briefly introduces the concept and some related studies of autoencoder. Some autoencoder variants from recent years are as follows: Stacked autoencoder (SAE), Stacked sparse autoencoder (SSAE), Locality constrained sparse autoencoder (LSAE), Weight-clustering sparse autoencoder (WCSAE), Distance constrained sparse stacking autoencoders (DCSSAE), Latent relationship guided stacked sparse autoencoder (LRSSAE), Stacked Denoising Sparse Autoencoders (SDSAE), Stacked Pruned Sparse Autoencoder (SPSAE).

## 3. Proposed Method

Fig. 3 shows the flowchart of the proposed NE_ESAE algorithm, which mainly includes neighboring sample envelope learning mechanism (NSELM), embedded stacked autoencoder (ESAE) and multilayer space ensemble mechanism (MSEM). The process of the NE_ESAE is described as follows.

1) The NSELM mines the spatial information of the samples to realize the multilayer transformation of samples.
2) ESAEs are trained in each layer of samples respectively to extract multilayer deep features.
3) The multilayer space ensemble mechanism (MSEM) fuses the classification results of the multilayer samples to realize the sample-feature cooperative transformation.

The main symbols used in this paper and their meanings are listed in Table 1.

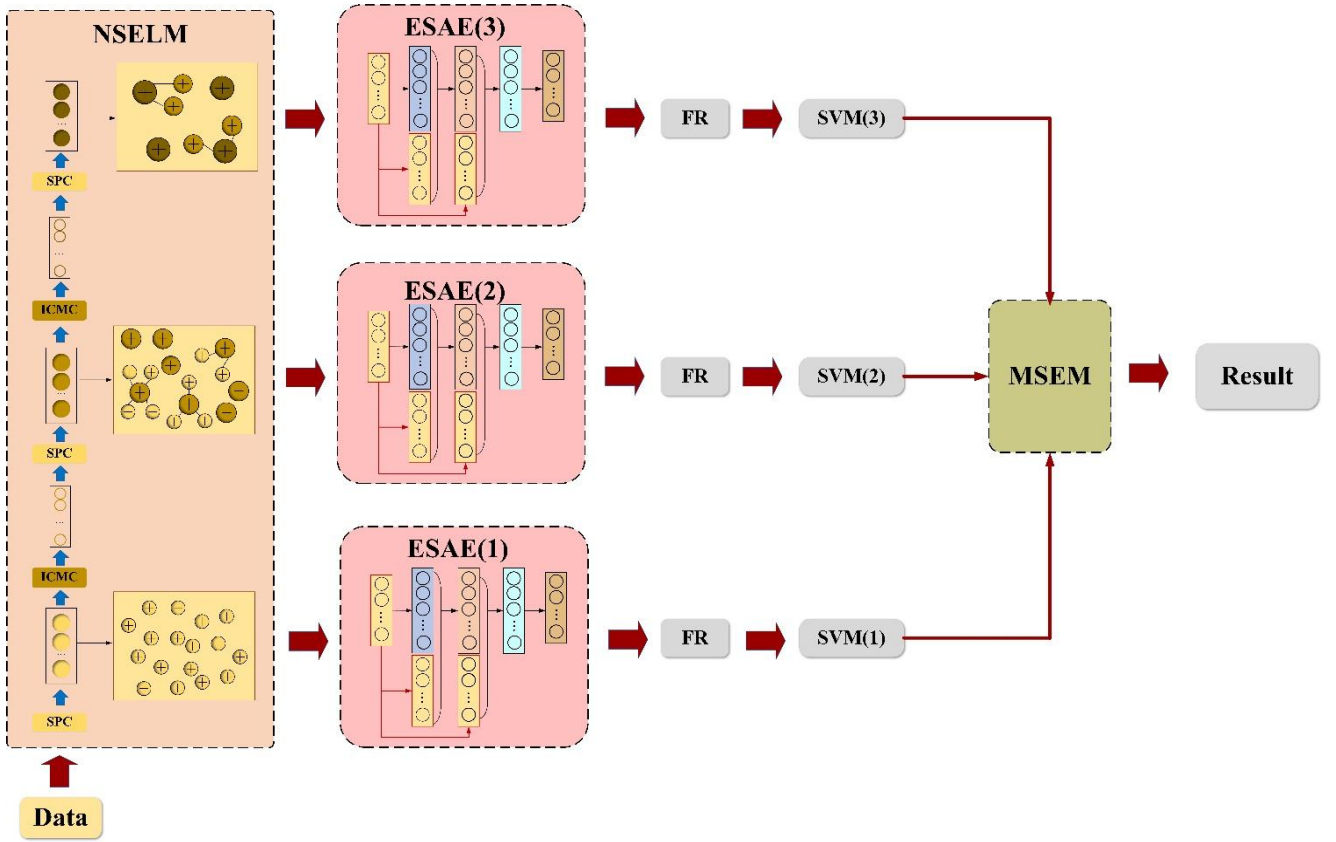

**Fig. 3** The proposed NS_ESAE model. Firstly, the original data is transformed by NSELM to obtain multilayer samples. Then the samples in each sample space are input ESAE for training to extract deep features. Finally, MSEM integrates multilayer samples to obtain the final classification results.

**Table 1** Main symbols and their meanings

| Symbol | Meaning |
| --- | --- |
| $\mathbf{X}$ | Original data, $\mathbf{X} \in \mathbf{R}^{n \times d}$ $n$ and $d$ are the original sample size and feature size respectively. |
| $x$ | The original sample. |
| $\phi(\cdot)$ | SPC operator, which is used to reconstruct the original sample. |
| $\mathbf{OS}$ | The original connection sample space. |
| $\mathbf{X}_{OS}$ | Generated sample pairs. $\mathbf{X}_{OS} \in \mathbf{R}^{n \times 2d}$ $n$ and $2d$ are the sample size and feature size, respectively. |
| $\mathbf{S}_P$ | Clustering sample set. |
| $\mathbf{PS}$ | The primary clustering sample space. |
| $\mathbf{SS}$ | The secondary clustering sample space. |
| $\mathbf{GIS}$ | The intermediate sample space. |
| $\mathbf{Q}$ | The generative transfer matrix. |
| $\varphi(\mathbf{X}_{PS})$ | The training set of the primary clustering sample. |
| $\varphi(\mathbf{X}_{OS})$ | The training set of the original connection sample. |
| $\varphi(\mathbf{X}_{GIS}^i)$ | The distribution of the generated intermediate sample. |
| $\varphi(\mathbf{X}_{OS}^{j})$ | The distribution of the original connection sample. |
| $\Psi$ | Linearly transformed projection matrix. |
| $\Theta$ | Auxiliary variable. |
| $\mathbf{E}$ | Augment Lagrangian multipliers. |
| $\rho 1$ | Penalty parameter. |
| $\mathbf{H}$ | Hidden feature, $\mathbf{H} \in \mathbf{R}^{n \times d^k}$. $d^k$ is the number of hidden-layer units of the kth AE. |
| $\lambda$ | Coefficient for the L2 weight regularization term. |
| $\beta$ | Coefficient for the sparsity regularization term. |
| $\rho$ | Sparse parameter. |
| $\hat{\rho}_j$ | Average activation value of all training samples on the $j$th hidden neuron. |
| $\mathbf{X}_{DF}$ | Data extracted by ESAE, $\mathbf{X}_{DF} \in \mathbf{R}^{n \times q}$. |
| $\mathbf{X}_{PF}$ | The matrix after the feature reduction. |

## 3.1 Neighbouring Sample Envelope Learning Mechanism (NSELM)

The proposed NSELM consists two parts. The first part is the sample pair connection (SPC) based on near-neighbor sample, which reconstructs sample pairs by connecting the original sample with the nearest-neighbor sample. The second part is ICMC, which can be seen as a combination of two-layer iterative mean clustering (IMC) and interlayer consistency mechanism (ICM). ICMC constructs a multilayer sample spaces, and achieves local and global consistency of structural information between interlayers, so as to better realize 'bigger' samples. The NSELM method is described in detail as follows.

## 3.1.1 Sample-pair connection based on near-neighbor (SPC)

In order to mine the neighborhood relationship between samples, this paper proposes SPC algorithm. We propose the $\phi(\cdot)$ operator for transforming (reconstructing) the original samples, which is to select a neighboring original sample, then capture the intrinsic structure of similar near-neighbor samples in the sample-pair connection space, and finally, connect the original sample with the near-neighbor sample to generate sample pair.

Given the original data $\mathbf{X}$, the neighboring sample is found as follows. First, we randomly pick $x_j$ from the same class sample set of $x_i$ in the $\mathbf{X}$, and calculate the Euclidean distance between $x_i$ and $x_j$. Based on this, we can find the neighboring sample $x_j'$, which are closest to the $x_i$ in terms of Euclidean distance. We use $\mathbf{D}$ to represent the distance, the objective function can then be expressed as

$$\min_{\mathbf{D}} \|x_i - x_j\|_2 \quad (1)$$

Therefore, concatenating $x_i$ and $x_j'$, the SPC method can be expressed as

$$\mathbf{X}_{OS} = \phi(x_i) = [x_i, x_j'] \quad (2)$$

The calculation process of $\phi(\cdot)$ is shown in Fig.4. In the figure, sample pairs $\mathbf{X}_{OS}$ are generated by SPC. We refer to the sample space consisting of $\mathbf{X}_{OS}$ as original sample pairs space (**OS**), which contains structural information among the original samples (relationship between neighboring original samples).

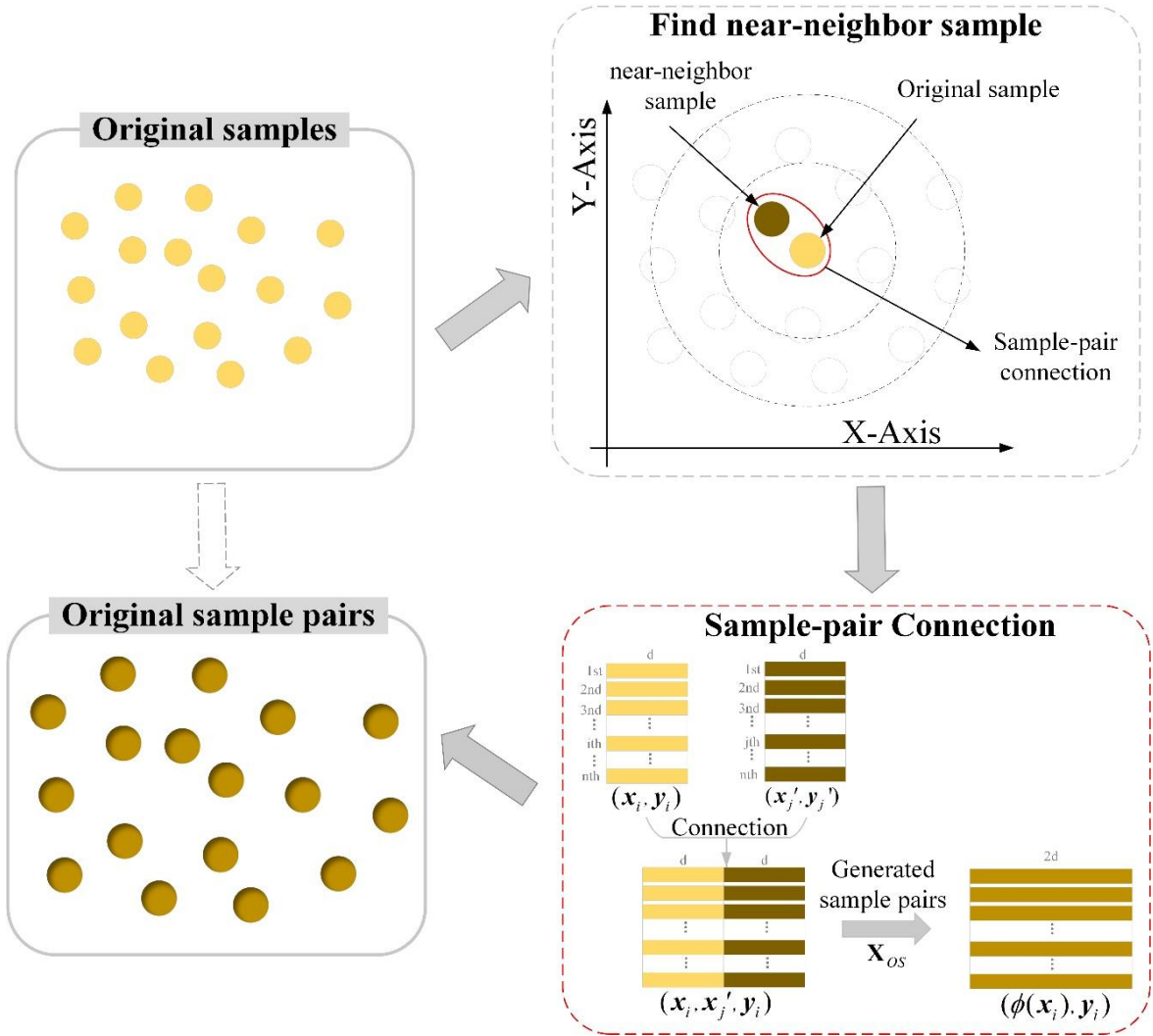

**Fig. 4** Sample pairs by connecting near-neighbor samples. First select an original sample; then capture the intrinsic structure of similar neighbor samples in the sample pair connection space; and finally connect the original sample and the near-neighbor sample to generate sample pairs.

### 3.1.2 Interlayer consistency mechanism based on multilayer iterative mean clustering (ICMC)

In order to mine the similarity relationship between samples, this paper proposes the ICMC. The proposed ICMC is shown in Fig. 5. As shown in the figure, ICMC mainly consists of IMC and ICM. IMCs is used to construct the multilayer sample spaces, and ICM is designed between interlayer, which can effectively reduce the difference in distribution between layers.

The proposed ICMC method is described in detail as follows.

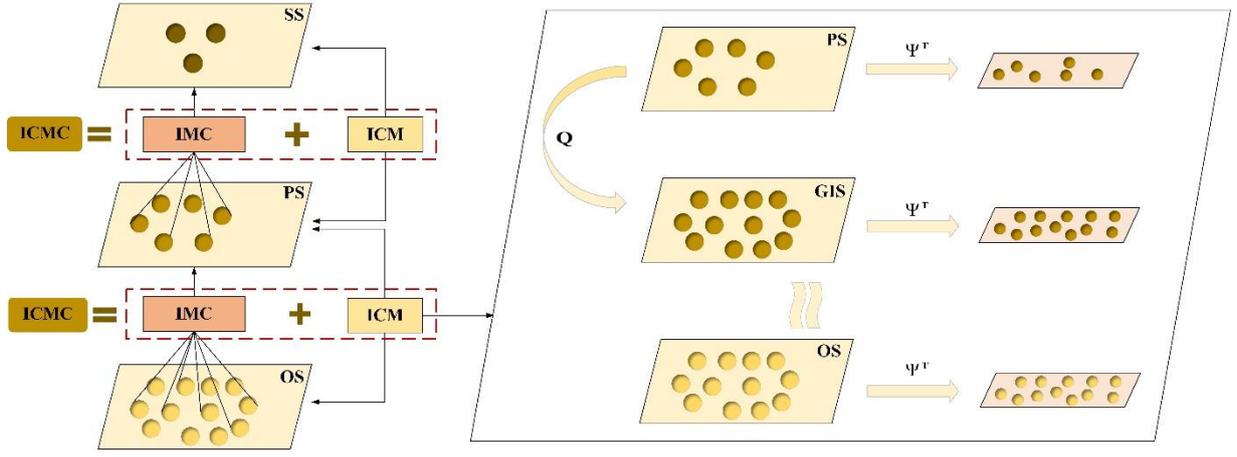

**Fig. 5** A flowchart of the proposed ICMC. ICMC mainly consists of IMC and ICM. IMCs is used to construct the multilayer sample spaces, and ICM is designed between interlayer, which can effectively reduce the difference in distribution between layers. In ICM, PS is used to generate GIS, and its sample distribution is similar to OS. The generation of GIS is carried out by the generation matrix $\mathbf{Q}$ in an unsupervised manner.

First, let $\mathbf{M} = [\mathbf{m}_1, \mathbf{m}_2, \cdots, \mathbf{m}_u]$ be the $u$ cluster centers, the formula of IMC is as follows

$$\min_{\mathbf{S}} \sum_{P=1}^{u} \sum_{\mathbf{Y} \in \mathbf{S}_P} \|\mathbf{Y} - \mathbf{M}_P\|^2 \tag{3}$$

where $\mathbf{S}_P$ is the sample set with $\mathbf{M}_u$ as the cluster center. To determine $\mathbf{S}_P$, the following two steps need to be repeated continuously:

Step 1: For each sample $\phi(\mathbf{x}_i)$, the Euclidean distance between $\phi(\mathbf{x}_i)$ and each cluster center is calculated in turn, and $\phi(\mathbf{x}_i)$ is divided into the cluster corresponding to the nearest sample center. Let the $p$th sample cluster generated in the $t$th clustering process be recorded as $\mathbf{S}_p^t$, and its allocation result can be expressed as

$$\mathbf{S}_p^t = \{\phi(\mathbf{x}_i) : \min_{1 \leq p \leq u} \|\phi(\mathbf{x}_i) - \phi(\mathbf{x}_p^t)\|^2\} \tag{4}$$

Step 2: After completing step 1, $u$ sample clusters are obtained, recalculate the sample center of each cluster by taking the mean value of all dimensions of all samples in the current cluster

$$\mathbf{M}_u^{t+1} = \frac{1}{card(\mathbf{S}_p^t)} \sum_{\phi(\mathbf{x}_i) \in \mathbf{S}_p^t} \phi(\mathbf{x}_i) \tag{5}$$

where $card(\cdot)$ is the number of samples in the $\mathbf{S}_P$. Repeat the above two steps until convergence, and take the final $u$ sample centers as a new sample set $\mathbf{X}_{PS} \in \mathbf{R}^{u \times 2d}$.

Based on the above steps, the primary clustering sample space (PS) can be obtained based on the original sample pairs (OS). By repeating the IMC process, we can obtain a secondary clustering sample space (SS). Therefore, the multilayer sample spaces and its respective samples constructed by ICMs are shown in Eq. (6). The OS is obtained by the SPC which the two neighboring samples are looked as an envelope; and the PS and SS are obtained by the SPC and IMC which the clustered samples are looked as an envelope. Therefore, the three layers of samples (OS, PS, and SS) are called three layers of envelope samples.

$$\text{Multilayer sample spaces} = \begin{cases} \text{OS, } \mathbf{X}_{OS} \in \mathbf{R}^{n \times 2d} \\ \text{PS, } \mathbf{X}_{PS} \in \mathbf{R}^{u \times 2d} \\ \text{SS, } \mathbf{X}_{SS} \in \mathbf{R}^{e \times 2d} \end{cases} \tag{6}$$

In order to ensure the consistency of sample distribution among multilayer sample spaces, we designed an ICM between neighboring sample spaces. Before this, we need to transpose the matrix in each sample space (i.e. $\mathbf{X}_{OS} \in \mathbf{R}^{2d \times n}$, $\mathbf{X}_{PS} \in \mathbf{R}^{2d \times u}$) to ensure the subsequent work. As shown in Fig.5, in ICM, we believe that PS can generate an intermediate sample space (GIS) with similar distribution to OS by generative transfer matrix $\mathbf{Q} \in \mathbf{R}^{u \times n}$. Our main purpose is to learn the generative transfer matrix $\mathbf{Q}$.

We use the implicit but generic transformation $\varphi$ to represent the training set of PS, OS and GIS, which are defined as $\varphi(\mathbf{X}_{PS})$, $\varphi(\mathbf{X}_{OS})$ and $\varphi(\mathbf{X}_{GIS})$. Where $\varphi(\mathbf{X}_{GIS}^i)$ and $\varphi(\mathbf{X}_{OS}^j)$ represents the distribution of the GIS and the OS, respectively, Considering that $\varphi(\mathbf{X}_{GIS}) = \varphi(\mathbf{X}_{PS})\mathbf{Q}$. Therefore, the ICM expression is

$$ICM(\text{GIS}, \text{OS}) = \frac{1}{n}\sum_{j=1}^{n}\left\|\varphi(\mathbf{X}_{GIS}^j) - \varphi(\mathbf{X}_{OS}^j)\right\|_2^2 + \sum_{i,j}^{n}\mathbf{W}_{ij}\left\|\varphi(\mathbf{X}_{GIS}^i) - \varphi(\mathbf{X}_{OS}^j)\right\|_2^2 + \left\|\mathbf{Q}\right\|_* \tag{7}$$

Overall, ICM consists of three parts. The first is the global distribution difference (GDD) loss, which minimizes the global difference in the edge distribution between GIS and OS. The second is local distribution difference (LDD) loss, which uses the locality of OS to measure the local area difference from GIS. The third term is LRC regularization to preserve the generalization of $\mathbf{Q}$.

## 3.2 Embedded Stacked Autoencoder (ESAE)

In order to mine the complementarity between the deep feature samples and the original feature samples, this paper proposed an embedded stacked autoencoder (ESAE). We designed an embedded unit on the basic SAE structure. The main function of this embedded unit is to introduce the input features into the training process of the encoder network, perform feature transformation through the embedding criterion, and use the obtained new features as the input data of the next layer. The ESAE model is shown in Fig. 6.

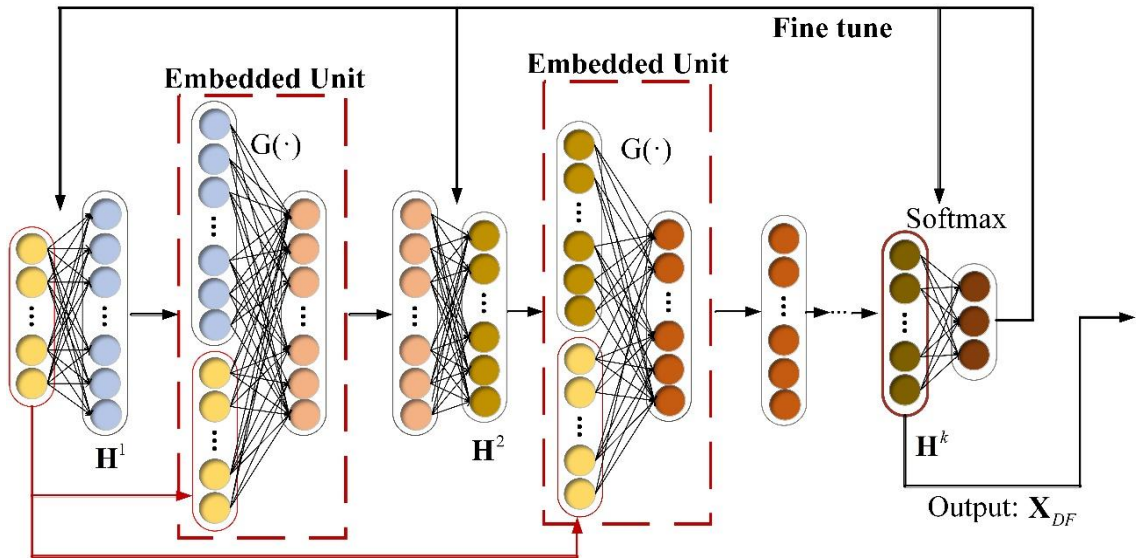

**Fig. 6** Structure of the proposed ESAE. The embedding unit introduces the original input feature information into the training process of the encoder network, performs feature transformation through embedding criteria, and takes the new feature as the input data of the next layer.

As shown in Fig. 6, ESAE consists of $K$ encoders cascaded together. Different from the traditional SAE, the key part of the ESAE is the embedding unit between two adjacent hidden layers. Assume that the input data of ESAE network is $\mathbf{X}_{OS}$, where the output matrix of the hidden layer in the $k$th encoder is $\mathbf{H}^k = [\mathbf{h}_1^k, \mathbf{h}_2^k, \cdots, \mathbf{h}_n^k] \in \mathbf{R}^{n \times d^k}$, $1 < k < K$, $d^k$ represents the number of hidden layer neurons of the $k$th encoder. In ESAE, the first layer of encoder training does not introduce an embedding unit, and its optimization goal is still to minimize the difference between the reconstructed data and the input data $\mathbf{X}_{OS}$. Starting from the second layer of the network, the encoded output of the previous encoder is not directly used as the input data of the next encoder, but first transformed by the embedding unit. The embedded unit can be expressed as

$$\mathbf{L}(\mathbf{V}) = \mathbf{G}^T(\mathbf{V}) \tag{8}$$

where $\mathbf{V} = \phi(x) \oplus h$, $\oplus$ means to concatenate the original input feature $\phi(x)$ and the hidden layer output feature $h$ of the encoder, $\mathbf{G}$ is the corresponding transformation matrix, consisting of 0 and 1. After transformation, the hidden layer output vector of the $k$th encoder in the ESAE network can be expressed as

$$h^k = f(\mathbf{W}_{k1}^T \mathbf{L}(\mathbf{V}^{k-1}) + \mathbf{b}_{k1}) \tag{9}$$

where $\mathbf{W}_{k1}$ and $\mathbf{b}_{k1}$ are the connection weight matrix and bias vector between the input layer and the hidden layer in the $k$th AE, respectively. And $f$ represents the activation function. The activation function of the encoder in this paper adopts the sigmoid function, which is $f(x) = 1/(1+\exp(-x))$. The purpose of the embedding unit is to introduce certain original information constraints in the layer-by-layer training process of the AE network, and filter some hidden layer outputs with weak class representation capabilities. Therefore, the objective function of the of the proposed embedding unit can be expressed as

$$\max_{\mathbf{G}} \ \mathrm{Tr}(\mathbf{G}^T \mathbf{V} \mathbf{V}^T \mathbf{G}) \\ s.t. \ \sum \mathbf{G}_{ij} = 2d \tag{10}$$

Calculate the covariance matrix $\mathbf{D}$ of the feature matrix $\mathbf{V}$ and sort its diagonal elements in descending order, take the first $2d$ values to form a vector $s \in \mathbf{R}^{2d}$. The transformation matrix $\mathbf{G}$ is obtained according to the following formula

$$\mathbf{G}_{ij} = \begin{cases} 1, & \text{if } s_j = \mathbf{D}_{ii} \\ 0, & \text{otherwise} \end{cases} \tag{11}$$

where $\mathbf{D}_{ii}$ is the $i$th diagonal element of the covariance matrix $\mathbf{D}$, $s_j$ is the $j$th element of the vector, and there is $i \in [1, n+2d], j \in [1, 2d]$. By combining Eq. (40)-(43), we can obtain the hidden layer output of the $k$th encoder. Then the decoding function of the $k$th encoder can be rewritten as

$$\mathbf{L}'(\mathbf{V}^{k-1}) = f(\mathbf{W}_{k2}^T h^k + \mathbf{b}_{k2}) \tag{12}$$

where $\mathbf{L}'(\mathbf{V}^{k-1})$ is the data obtained by reconstructing the output of the embedding unit. $\mathbf{W}_{k2}$ and $\mathbf{b}_{k2}$ are the connection weight matrix and bias vector between the hidden and output layers in the kth encoder, respectively.

In order to achieve an "overcomplete" nonlinear mapping of the input vector, sparse criterion can be imposed in the hidden layers of ESAE. We introduce KL divergence to enable the hidden layer of the encoder to learn the sparse representation. The formula for KL divergence is expressed as follows

$$\mathrm{KL}(\rho \| \widehat{\rho}_j) = \sum_{j=1}^{2d} \left( \rho \log(\frac{\rho}{\widehat{\rho}_j}) + (1-\rho) \log(\frac{1-\rho}{1-\widehat{\rho}_j}) \right) \\ \widehat{\rho}_j = \frac{1}{2d} \sum_{1}^{2d} f^j(x^i) \tag{13}$$

where $f^j(x^i)$ is the activation value of the $i$th input vector on the $j$th neuron in the hidden layer.

After introducing the embedding unit in the structure and introducing the sparse criterion in the training process, the optimization objective function of the $k$th encoder of ESAE is written as

$$\arg\min_{\theta} \frac{1}{n} \sum_{i=1}^{n} \left\| \mathbf{L}(\mathbf{V}^{k-1}) - \mathbf{L}'(\mathbf{V}^{k-1}) \right\|^2 + \lambda(\|\mathbf{W}_{k1}\|_2 + \|\mathbf{W}_{k2}\|_2) + \beta(\sum_{j=1}^{d^k} \mathrm{KL}(\rho \| \widehat{\rho}_j)) \tag{14}$$

where $\lambda$ and $\beta$ are the penalty parameters of the regularization item and the sparse criterion item, respectively. The training with Eq. (47) as the objective function is the pre-training stage of each layer in the ESAE network. After that, the pre-trained hidden layers are cascaded to form a deep network model, and the model parameters obtained from the pre-training will be used as the deep network. The pre-training stage is an unsupervised learning process. In order to obtain features with stronger classification ability, discriminative information is introduced into the network, that is, a softmax layer is connected to the hidden layer of the last encoder as a classification layer, and the classification layer is supervised in a supervised way.

The proposed ESAE not only takes advantage of the feature that deep encoder networks can automatically learn latent relationships between data, but also improves by introducing initial information and category information into network training, this improved the robustness of deep features.

After two stages of network pre-training and fine-tuning, for the $i$th input vector $\mathbf{X}_{iOS} = [\phi(\mathbf{x}_{i1}), \phi(\mathbf{x}_{i2}), \cdots, \phi(\mathbf{x}_{i2d})]$, each hidden layer in the network outputs a new feature vector, representing different levels of information. We take the output of the last hidden layer as the deep feature learned by the network, which can be expressed as follows

$$\mathbf{X}_{DF} = [\mathbf{x}_1', \mathbf{x}_2', \cdots, \mathbf{x}_d'] \in \mathbf{R}^{n \times q} \tag{15}$$

where $q$ is the number of neurons in the last hidden layer.

## 3.3 Multilayer Space Ensemble Mechanism (MSEM)

In order to improve the complementarity of multilayer sample space, this paper proposes the MSEM. MSEM fuses the ESAE models under each layer of sample space to achieve final result. This MSEM will discuss two commonly used ensemble methods: weighted fusion method (WF) and majority voting method (MV). By NSELM, our model contains three layers of sample spaces, which are OS, PS and SS, which contain structure information among samples. Suppose that the prediction result of the model in the OS is $y_0$, the prediction result of the model in the PS is $y_1$, the prediction result of the model in the SS is $y_2$, the realization process of multilayer space ensemble is as follows.

### 3.3.1 Weighted fusion method (WF)

The key to WF is to determine the weight $\mathbf{w}_i$ of the classifier model in each sample space, which requires the use of a validation set. Assuming that the prediction result of each encoder on the validation set is $\mathbf{r} = (\mathbf{r}_0, \mathbf{r}_1, \mathbf{r}_2)$, The optimal weight vector is $\mathbf{w}_f = (\mathbf{w}_0, \mathbf{w}_1, \mathbf{w}_2)$, The optimal weight vector can be obtained by the following formula:

$$\arg\max_{\mathbf{w}_f} = \sum_{i=0}^{\Lambda_v} \eta(round(\mathbf{w}_i^T \mathbf{r}_i), y_i) \tag{16}$$
$$s.t. \mathbf{w}^T \mathbf{I} = 1, \mathbf{I} = (1, 1, ..., 1)$$

where $\Lambda_v$ is the number of validation set, $\eta(a,b) = \begin{cases} 1, a = b \\ 0, a \neq b \end{cases}$, let $y = (y_0, y_1, y_2)$. The prediction result of the final test sample can be expressed as

$$y_F = round(\mathbf{w}_f^T y) \tag{17}$$

### 3.3.2 Majority vote method (MV)

Assuming that the dataset has a total of $C$ categories, for each prediction sample, a vector for statistics is given, and the initial value is

$$count = (n_i, n_j, ..., n_C), n_j = \big|_{j=t}^{C} \tag{18}$$

The prediction result of the classifier model under each sample space is $y_i \begin{cases} n \\ i=0 \end{cases}$, then

$$n_j = \big|_{j=t}^{C} = \begin{cases} n_j + 1, & \text{if } y_i = j \\ n_j, & \text{others} \end{cases}, j = 1, \cdots, n \tag{19}$$

The final classification label $y_F$ is the subscript $j$ corresponding to the largest component in the statistical vector $count$, namely

$$y_F = j \tag{20}$$
$$s.t. n_j = \max(count)$$

The NE_ESAE proposed in this paper is summarized in Algorithm 1.

**Algorithm 1** NE_ESAE

**Input:** Data matrix $\mathbf{X}$

**Procedure:**

1. Get $\mathbf{X}_{OS}$ according to SPC method
2. Construct multilayer sample spaces OS, PS, SS based on IMCs
3. Construct a new sample $\mathbf{X}_{PS}'$, $\mathbf{X}_{SS}'$ based on ICM
4. **For** $i = 0,1,2$ **do**
   4.1 Train ESAE network in sample space $\mathbf{X}_i'$ and extract deep features
   4.2 Get the $\mathbf{X}_{DF}$ in the current sample space
**End for**
5. Obtain the new data matrix $\mathbf{X}_{PF}$ according to FR
6. Fusing classification results from multilayer sample spaces
   6.1 Determine $\mathbf{w}_f$ based on the validation set and Eq. (16), and obtain the final label according to Eq. (17)
   6.2 Calculate the statistical vector *count* according to Eq. (18), and obtain the final label according to Eq. (20)

**Output:** Predicted label $y_F$

## 4. Experimental results and analysis

In order to verify the effectiveness of proposed NE_ESAE, four groups of experiments are conducted in this paper. The first group of experiments based on ablation study is to verify the effectiveness of SPC and ICMC, the effectiveness of MSEM, and the effectiveness of NSELM and ESAE respectively. The second group of experiments compares the proposed algorithm with existing representative feature learning algorithms and other stack representative SAE algorithms. The third group of experiments analyzes the effects of some parameters including the type of classifier and the proportion of cluster center samples. The fourth group is about confusion matrix experiments.

## 4.1 Experimental conditions

The performance of the proposed algorithm is tested on 12 datasets with different sample sizes and feature dimensions. These datasets are representative. They have both large and small samples, binary and multiple class labels, and high, medium and low dimensional features. The main information (including the size of samples, number of features, number of categories, and relevant paper) of all dataset is shown in Table 2.

Table 2 Basic information of datasets used in study

| Dataset | Instances | Attributes | Class | Relevant paper |
|---|---|---|---|---|
| Statlog_Landsat_Satellite （Statlog） | 6435 | 36 | 6 | Reference [63] |
| Pen-Based Recognition of Handwritten Digits（Pendigits） | 10992 | 16 | 10 | Reference [64] |
| LSVT Voice Rehabilitation Dataset (LSVT) | 126 | 309 | 2 | Reference [65] |
| Urban land cover (Urban) | 675 | 148 | 9 | Reference [66] |
| Statlog Heart Dataset (Heart) | 270 | 13 | 2 | Reference [67] |
| Pima Indians Diabetes Dataset (PID) | 768 | 8 | 2 | Reference [68] |
| Parkinson Speech Dataset (PD) | 1040 | 26 | 2 | Reference [69] |
| Alzheimer's disease (AD) | 90 | 32 | 3 | Reference [70] |
| Breast Cancer Wisconsin Original (Wisconsin) | 683 | 9 | 2 | Reference [71] |
| Maxlettle Parkinson Dataset (Maxlettle) | 195 | 22 | 2 | Reference [72] |

| Statlog_Vehicle Silhouettes (Vehicle) | 846 | 18 | 4 | Reference [63] |
| Breast Cancer Wisconsin Diagnostic (WDBC) | 569 | 30 | 2 | Reference [73] |

In the following experiments, the number of layers of the encoder in the proposed ESAE network is set to 3; the number of hidden layer neurons of each encoder is determined according to the number of samples and feature dimensions of the dataset; and the optimal structure is determined by grid search method.

The experiment adopts the hold-out cross-validation method, and each dataset is divided into three equal parts. 1/3 of the samples are for training set, 1/3 of the samples are for validation set, and 1/3 of the samples are for test set. The classification model used in the experiment is SVM. In order to eliminate the influence of randomness, each experiment was repeated five times to obtain the average and variance (stand deviation) to characterize the classification effect of the algorithm in this paper.

## 4.2 Algorithm comparison-Compared with the representative stack autoencoders

To verify the advantage of NE_ESAE over existing SAEs, some representative SAEs are considered for comparison. They include the stacked autoencoder (SAE), stacked sparse autoencoder (SSAE) [52], stacked denoising autoencoder (SDSAE) [57], stacked pruning sparse autoencoder (SPSAE) [58], and embedded stacked group sparse autoencoder ensemble with L1 regularization and manifold reduction (ESGSAE_FF) [74]. Because some SAEs' study do not disclose their source code and corresponding experimental results in the paper, we choose the methods that have the corresponding experimental results reported in the published papers for comparison. The experimental results are presented in Table 3.

Table 3 Classification accuracy (mean ±variance) of different deep autoencoder classifiers

| Dataset | SAE (%) | SSAE (%) | SDSAE (%) | SPSAE (%) | ESGSAE_FF (%) | **NE_ESAE (%)** |
|---|---|---|---|---|---|---|
| AD | 50.67±7.95 | 56.67±5.27 | 55.58±4.36 | 57.78±4.27 | 67.33±2.49 | **90.67±6.41** |
| LSVT | 83.80±5.16 | 83.33±5.83 | 76.62±5.29 | 84.33±5.36 | 92.76±0.62 | **96.67±3.61** |
| PD | 61.15±2.91 | 64.48±2.05 | 64.88±1.84 | 64.22±2.34 | 66.72±0.87 | **78.74±4.22** |
| Urban | 74.48±3.33 | 79.73±0.67 | 75.17±1.88 | 77.81±1.17 | 83.20±1.01 | **84.29±3.38** |
| Vehicle | 67.30±3.33 | 70.00±2.99 | 72.00±2.25 | 74.76±2.93 | 81.91±0.42 | **89.53±1.96** |
| pendigits | 89.64± 1.44 | 93.80± 0.51 | 94.58 ± 0.53 | 91.60 ± 0.57 | **98.00 ±0.12** | 84.21±0.83 |
| Statlog | 83.67± 0.36 | 84.85± 0.84 | 83.65± 0.71 | 85.87± 0.86 | 87.28±0.12 | **87.93±0.93** |

As the Table 2 shows, proposed method NE_ESAE has the best classification accuracy compared with other stack autoencoder algorithm in most cases. For example, the classification accuracy of NE_ESAE in AD dataset is 40.00% higher than SAE, 34.00% higher than SSAE, 35.09% higher than SDSAE, 32.89% higher than SPSAE, and 23.34% higher than ESGSAE_FF, respectively. The experimental results prove that it is feasible to consider the sample structure information and introduce the complementary ideas of original features and deep features in pre-training. For datasets with small number of samples and high feature such as LSVT and Urban, the NE_ESAE model still shows better performance. The one of the possible reasons is that the NSELM considers the latent data structure information among samples, so that the neglected intrinsic information among samples can be better learned. The experimental results demonstrate that proposed NE_ESAE is an efficient new SAE which have better feature learning ability.

## 5. Conclusion

Feature learning is very important in machine learning algorithms. SAE is a very important algorithm for feature learning. It has unique and excellent performance. However, existing deep SAEs only focus on original samples without considering the hierarchical structure between samples. In order to solve this problem, NE_ESAE ensemble model is proposed. Firstly, the

NSELM is proposed for preprocessing of input of SAE. NSELM constructs sample pairs and a multilayer sample spaces, which considers the hierarchical structure between samples and generates layers of envelope samples with better quality samples; Secondly, ESAE is proposed and trained in each layer of sample space to consider the original samples during training and in the network structure, thereby better finding the relationship between original feature samples and deep feature samples. Thirdly, feature reduction and base classifiers are conducted on the layers of envelope samples respectively, and output classification results of layers samples. Finally, the classification results of the layers of envelope sample space are fused through the MSEM to realize the final results.

To verify the effectiveness of NE_ESAE, we experimentally compared its performance with some representative SAEs, including SSAE, SAE, SPSAE, SDAE and ESGSAE_FF. The results show that our proposed method has obvious advantages over most algorithms. Taking AD dataset as an example, the classification accuracy of this method is 40.00%, 34.00%, 35.09%, 32.89%, 23.34% higher than that of other algorithms, respectively. Finally, the influence of the classifier type and the cluster ratio in the IMC on our method is analyzed, which provides a certain reference for parameter selection.

The main contribution of this paper is that proposed the NSELM for the first time. This method not only deeply mines the information of the data, but also learns the structural information of each layer of samples, which increases the complementarity of the features and enhances the representation ability of the features. The ESAE designed in this paper is a lightweight deep network. The network introduces the original features into the structure and training process of the autoencoder, which improves the complementarity between the deep features and the original features. Able to achieve high-quality feature learning. Based on the NSELM and ESAE, NE_ESAE model is proposed to realize the sample-feature cooperative transformation. As can be seen from further analysis, the proposed NE_ESAE model has the significant advantages of better generalization ability, more stable and general model, and higher classification accuracy. However, the method proposed in this paper also has some limitations. First of all, NE_ESAE mainly focuses on improving feature representation learning ability, and does not consider other types of data, such as image data and speech data. Second, the method is not strictly end-to-end. In the future, we plan to extend the proposed approach in two directions. One is to explore network structure to enhance feature learning ability, and the other is to incorporate more types of data.

**Acknowledgements**

**Data availability**
The data and codes can be found in: https://github.com/cutezoe00/NEESAE

**Conflict of interest**
The authors declare no conflicts of interest pertaining to this work.

**Ethics Approval and Consent to Participate**
Not applicable.

**Consent for publication**
Not applicable.